\documentclass[a4paper, 10 pt, conference]{ieeeconf}  
\makeatletter
\def\endthebibliography{%
	\def\@noitemerr{\@latex@warning{Empty `thebibliography' environment}}%
	\endlist
}
\makeatother

\IEEEoverridecommandlockouts                              

\usepackage{cite}
\usepackage{amsmath,amssymb,amsfonts}
\usepackage{algorithmic}
\usepackage{graphicx}
\usepackage{textcomp}
\usepackage{xcolor}
\usepackage{siunitx}
\usepackage{tikz}
\usepackage{pgfplots}

\newcommand\copyrighttext{%
	\footnotesize \textcopyright 2020 IEEE. Personal use of this material is permitted.
	Permission from IEEE must be obtained for all other uses, in any current or future
	media, including reprinting/republishing this material for advertising or promotional
	purposes, creating new collective works, for resale or redistribution to servers or
	lists, or reuse of any copyrighted component of this work in other works.}
\newcommand\copyrightnotice{%
	\begin{tikzpicture}[remember picture,overlay]
	\node[anchor=south,xshift=10pt,yshift=10pt] at (current page.south) {\fbox{\parbox{\dimexpr\textwidth-\fboxsep-\fboxrule\relax}{\copyrighttext}}};
	\end{tikzpicture}%
}

\title{\LARGE \bf
Radar-based Dynamic Occupancy Grid Mapping and Object Detection
}

\author{Christopher Diehl$^{1}$, Eduard Feicho$^{2}$, Alexander Schwambach$^{2}$, Thomas Dammeier$^{2}$,\\ Eric Mares$^{1}$ and Torsten Bertram$^{1}$
\thanks{$^{1}$ Christopher Diehl and Torsten Bertram are, and Eric Mares was with the Institute of Control Theory and Systems Engineering, TU Dortmund, 44227 Dortmund, Germany
        {\tt\small forename.surename@tu-dortmund.de}}%
\thanks{$^{2}$ Eduard Feicho, Alexander Schwambach and Thomas Dammeier are with HELLA Aglaia Mobile Vision GmbH, 12109 Berlin, Germany
        {\tt\small forename.surename@hella.com}}%
}

\begin{document}

\maketitle
\thispagestyle{empty}
\pagestyle{empty}

\copyrightnotice
\begin{abstract}
Environment modeling utilizing sensor data fusion and object tracking is crucial for safe automated driving. In recent years, the classical occupancy grid map approach, which assumes a static environment, has been extended to dynamic occupancy grid maps, which maintain the possibility of a low-level data fusion while also estimating the position and velocity distribution of the dynamic local environment. 
This paper presents the further development of a previous approach. To the best of the author's knowledge, there is no publication about dynamic occupancy grid mapping with subsequent analysis based only on radar data. Therefore in this work, the data of multiple radar sensors are fused, and a grid-based object tracking and mapping method is applied. Subsequently, the clustering of dynamic areas provides high-level object information. For comparison, also a lidar-based method is developed. The approach is evaluated qualitatively and quantitatively with real-world data from a moving vehicle in urban environments. The evaluation illustrates the advantages of the radar-based dynamic occupancy grid map, considering different comparison metrics. 

\end{abstract}

\section{Introduction}
Sensor fusion can be done on different levels of information processing. Often detection of objects is performed per sensor, and then the hypotheses are fused on a high-level abstraction like in \cite{Aeberhard2017}. Occupancy grids, introduced in \cite{Elfes1989}, provide a possibility for low-level data fusion with less information loss and solve the data association problem implicitly by a spatial mapping into a cell-discretized map. Also, they can represent objects of arbitrary shapes but are constructed under the assumption of a static environment. This leads to artifacts in the presence of dynamic objects in the local environment. Therefore, previous approaches combine the estimation of static and dynamic areas using a grid-based environment representation. The resulting dynamic occupancy grid map (DOGM) contains information about the occupancy and the velocity in each grid cell. 
Most publications use lidar or camera sensors as a primary information source for constructing DOGM. Radar sensors, which are cheaper but noisier compared to lidar sensors, are only used as an addition. 
This work presents as first an approach with subsequent analysis based only on radar data.

 There are three main contributions. First, the method described in \cite{Tanzmeister2017} is further developed concerning utilizing only radar data. Second, a clustering algorithm detects dynamic areas. Third, the usage of radar-only based DOGMs is analyzed through comparison with a lidar-only based approach regarding different evaluation metrics.

The remainder of the paper is organized as follows. Section II summarizes related publications.
Section III introduces the problem of grid-based object tracking, evidential mapping, and dynamic object information extraction. The
developed radar-based method is evaluated in Section IV and compared with a lidar-based one under different comparison metrics. Section V concludes the paper and gives a further outlook.

\section{Related Work}

The DOGM contains velocity and occupancy information. The basis for one group of approaches is the Bayesian Occupancy Filter (BOF) \cite{Coue2006}, which recursively estimates the underlying probability distribution utilizing a four-dimensional grid, i.e., two dimensions each for the position and the speed. This contribution comes with computational issues and the need for velocity discretization.  Therefore, the Sequential Monte Carlo Bayesian Occupancy Filter (SMC-BOF) \cite{Danescu2011} is introduced. A particle filter estimates the underlying position and velocity distribution of the grid based on stereo-images. Tanzmeister et al. \cite{Tanzmeister2014} adopt this approach and use data of multi-layer laser scanners. These methods can misinterpret elongated static obstacle structures like guardrails as dynamic objects. Therefore, other contributions use additional radar Doppler information to reduce this effect \cite{Tanzmeister2017}, \cite{Nuss2014}, \cite{Steyer2018a}. Moreover, radar measurements improve the dynamic estimation and lead to faster convergence of the filter. The approach of Nuss et al. \cite{Nuss2014} has been evolved in \cite{Nuss2017} while using methods of the random finite set theory and an efficient GPU implementation. Vatavu et al. fuse camera, lidar, and radar data to a multi-channel grid that is used in the particle filter solution \cite{Vatavu2018}. Using semantic information improves the accuracy of the grid estimation.
Recently a recurrent neural network-based contribution for DOGM mapping utilizing lidar sensors has been published in \cite{RNNBosch}. It outperforms the approach presented in \cite{Nuss2017}, but the authors only tested it on a stationary car.  

 For applications such as motion planning or situation analysis, an abstraction of the dynamic environment on the object level is desired. Therefore different techniques have been developed to extract object information from grid maps. On the one hand, there are deep learning-based methods (e.g. \cite{Engel2018}) depending on manually labeled data.  The other group of approaches uses clustering algorithms, which depend on hand-crafted thresholds \cite{Steyer2017},\cite{Gies2018}.

\section{Radar-based Dynamic Occupancy Grid Mapping}
This section describes the technical details of the developed method for constructing a DOGM. It is built upon a grid-based tracking and evidential mapping approach presented in \cite{Tanzmeister2017}, which estimates an evidential map $\textbf{M}_{1:t}$ based on a measurement grid $\textbf{M}_{t,z}$. Moreover, the first two statistical moments per cell are estimated, considering the underlying particle population. The concatenation of the channels results in the DOGM $\textbf{G}_{1:t}$. Subsequently, clustering extracts dynamic object information.


\subsection{Environment Representation}
The goal is to derive occupancy hypotheses of static, dynamic, free space areas, and velocity information in a map from past sensor measurements $\textbf{z}_{1:t}$ up to time $t$. Therefore a grid-based environment model is used, where every grid cell contains evidence masses of the Dempster-Shafer theory \cite{Dempster},\cite{Shafer}. The frame of discernment is chosen as 
\begin{equation}
\Omega=\{F,S,D\},  
\label{FrameOfDiscrementDOG}
\end{equation} whereby the considered individual hypotheses of the power set $2^\Omega$ are the free space $\{F\}$, occupancy from static obstacles $\{S\}$, occupancy from dynamic obstacles $\{D\}$, occupancy from either static or dynamic obstacles $\{S,D\}$, and the residual uncertainty $\{\Omega\}$
 as proposed in \cite{Tanzmeister2017}. Thereby $m(A)$ describes the evidence for the hypotheses $A \in 2^{\Omega}$. Moreover, our DOGM contains channels that describe the first two stochastic moments. Then the set $\Theta$ defines the channels of $\textbf{G}_{1:t}$ with
\begin{multline}
\Theta=\{m(F),m(S),m(D),m(\{S,D\}),m(\Omega), v_x, v_y, \\ \sigma^2_{v_x}, \sigma^2_{v_y}, \sigma_{v_x,v_y} \}, 
\label{ChannelDOG}
\end{multline}
where $\sigma^2_{v_x}$, $\sigma^2_{v_y}$ are the variance of the estimated velocity in $x$- and $y$-direction and $\sigma_{v_x,v_y}$ is the covariance .
The evidence masses $m(M_{1:t}^i)$ with $M_{1:t}^i \subseteq \{F,S,D\}$ for every grid cell $i \in \{1, \dots, I\}$ of the map $\textbf{M}_{1:t}$ with $I$ cells and the stochastic moments are estimated based on the joint track variable 
\begin{equation}
\textbf{x}=[\textbf{s}~\textbf{v}]^T=[ s_x ~ s_y ~ v_x ~ v_y]^T.
\label{joint track variabel}
\end{equation} The 4-D vector $\textbf{x}$ contains the 2-D position $\textbf{s}$ and the 2-D velocity $\textbf{v}$. Let $\textbf{q}_t$ be the ego-pose consisting of a 2-D position and an orientation at time $t$. Then the goal is to estimate the joint belief distribution 
\begin{equation}
m\big(\textbf{M}_{1:t},\textbf{x}_t|\textbf{z}_{1:t},\textbf{q}_{1:t}\big)=p\big(\textbf{x}_t|\textbf{z}_{1:t}, \textbf{q}_{1:t}\big) m\big(\textbf{M}_{1:t}|\textbf{x}_{1:t}\textbf{z}_{1:t},\textbf{q}_{1:t}\big).
\label{FactorizationEvidentialMappingTracking}
\end{equation}  This can be factorized into a probabilistic tracking problem $p\big(\textbf{x}_t|\textbf{z}_{1:t}, \textbf{q}_{1:t}\big)$ and a conditional evidential mapping problem $m\big(\textbf{M}_{1:t}|\textbf{x}_{1:t},\textbf{z}_{1:t},\textbf{q}_{1:t}\big)$.  

\subsection{Measurement Grid}
First, the current sensor measurements $\textbf{z}_t$ are transferred into the grid-based representation $\textbf{M}_{t,z}$. Therefore, an inverse sensor model inspired by the work presented in \cite{Homm2010} is applied. Occupancy probabilities of the cells corresponding with a measurement are modeled as a 2-D gaussian. 
In addition to that, the free space probability decreases with increasing distance under the assumption that small objects are detected less frequently far away. Cells behind a measurement are modeled as uncertain. The methodology described in \cite{Moras2011} is adopted to obtain evidence masses from occupancy probabilities in the frame of discernment (\ref{FrameOfDiscrementDOG}). One cell $i$ of the resulting measurement grid contains masses $m_{t,z}(\{S,D\}^i)$, $m_{t,z}(F^i)$, and $m_{t,z}(\Omega^i)$. This work uses a multi-sensor-system. Therefore, a measurement grid is computed per sensor and fused through the use of Dempster's rule

\begin{equation}
m_{1 \oplus 2}(A) = \frac{\sum_{B \cap C = A} m_1(B)m_2(C)}{1-\sum_{B \cap C = \emptyset} m_1(B)m_2(C)} 
\label{DS-Kombinationsregel}
\end{equation} $\forall A,B,C \subseteq \Omega \neq \emptyset $.
For comparison, also a measurement grid based on lidar sensors with a lower uncertainty in the used inverse sensor model is constructed.

\subsection{Grid-based Tracking}
This section describes the grid-based tracking approach inspired by the work of \cite{Tanzmeister2017}. The probabilistic tracking problem is formulated as an $I$-mixture model, where every mixture represents a cell $i$ and is approximated by a set of particles. The resulting filtering distribution can be described with
\begin{equation}
p\big(\textbf{x}_t|\textbf{z}_{1:t}, \textbf{q}_{1:t}\big)\approx \sum_{i=1}^{I} \lambda^i_t \sum_{k\in K^i_t} w_{t,[k]} \delta\big(\textbf{x}_t;\textbf{x}_{t,[k]}\big), 
\label{Filterverteilung}
\end{equation} 
where $k$ denotes the particle index from the set of particles $K^i_t$ at time $t$, and $\delta(\cdot;\textbf{x})$ describes the Dirac distribution at $\textbf{x}$. The method uses $I$ particle filters with individual particle weights $w_{t,[k]}$. The particles can move between the cells, which have a mixture weight, here also referred to as cell weight, $\lambda^i$. The low-level tracking consists of four main components.  The following sections will introduce them briefly and will describe our changes compared to the original approach.
\subsubsection{Sampling}
First of all, sampling of particles from the proposal distribution is performed in every cell $i$, as described in \cite{Tanzmeister2017}. It is a mixture of the prior distribution, which is built upon the predicted previous particle population, and the distribution based on the current measurement $\textbf{z}_t$. The algorithm samples the position and velocity component of newly generated particles from the latter distribution.
In the adapted method, $w_s(z^i_{v_\textrm{R}})$ describes the amount of sampled static particles and is a function of the measured absolute radial velocity $z^i_{v_\textrm{R}}$.
 This leads to errors in the case of objects that move orthogonally to the sensor coordinate system. Because then, no radial velocity component is measured, and the algorithm samples a wrong particle.
In this work, $w_s(b_{\textrm{RSP}})$ is a function of the variable $b_{\textrm{RSP}}$. It pre-classifies a radar measurement, whether it is static or dynamic. Therefore, different properties of the radar raw signal, such as direction, signal-to-noise ratio, measured radial velocity, and vehicle ego-motion, are used. If $b_{\textrm{RSP}}$ is below a threshold $\epsilon_{\textrm{RSP}}$, then no static evidence exists, and dynamic particles are sampled. Otherwise, the algorithm generates static particles.  

\subsubsection{Weighting of Particles}
After sampling, the particles are weighted with the help of the mixture distribution, as described in \cite{Tanzmeister2017}.
Therefore, particles with states which fit the actual measurement receive a high unnormalized weight $\overline{w}_{t,[k]}$.

\subsubsection{Weighting of Mixtures}
The next step is the weighting of the individual mixture components. The unnormalized mixture weight $\widetilde{\lambda}^i_t=$ regulates the number of surviving particles in a cell. It is determined based on the maximum value of the current occupancy mass ${m}_{t,z}(\{S,D\}^i)$ in cell $i$ of the measurement grid $\textbf{M}_{t,z}$  and a used decay function  $f_{\textrm{dec}}$, as described in \cite{Tanzmeister2017}. Because of missing information about  $f_{\textrm{dec}}$ in the original approach, an exponential decay function based on \cite{Steyer2018a} is introduced
\begin{equation}
f_{\textrm{dec}}\big(\overline{\lambda^i_t},m_{t,z}\big(F^i\big)\big)=k_\textrm{d} \min\big(\overline{\lambda^i_t},1\big)\big(1-m_{t,z}\big(F^i\big)\big).
\end{equation} 
Let $\overline{\lambda^i_t}$ be the predicted unnormalized mixture weight, which equals the sum of previous normalized weights of particles, which are now in cell $i$.
The decay function reduces the cell weight concerning the parameter $k_\textrm{d}$. If no new measurement is observed, particles should not be destroyed immediately because noisy and sparse radar data could have caused the missed detection. Therefore, the robustness is increased.
\subsubsection{Resampling}
Lastly, the particles are resampled based on their weight. Therefore, particles with a high weight have a high survival probability in the cell. 
\subsection{Evidential Mapping}
The result of the grid-based tracking is a weighted particle population, which approximates the position and velocity distribution of the vehicle environment. Based on this, an evidential grid map $\textbf{M}_{1:t}$ is created. Therefore, the mass distribution $m\big(\textbf{M}_{1:t}|\textbf{x}_{1:t}, \textbf{z}_{1:t} \textbf{q}_{1:t})$ over the map consisting of $I$ sets $M^i_{1:t}$ of evidence variables is estimated. The equivalent to the measurement grid in the classical occupancy grid approach is the map $\textbf{M}_t$, whose evidence masses are estimated based on the particle population at time $t$. First, particles are classified based on their age and velocity. The ones who reach an age $a_{\textrm{min}}$ are classified as static if their velocity $v_{t,[k]}$ is below a threshold $||v_{t,[k]}|| \leq \epsilon_v$. Otherwise, the algorithm classified them as dynamic. Each particle with age $a_{t,[k]} < a_{\textrm{min}}$ remains unclassified. Then evidence masses from the frame of discernment (\ref{FrameOfDiscrementDOG}) are derived as in \cite{Tanzmeister2017}. Afterward, the old map $\textbf{M}_{1:t-1}$ is updated with the modified map $\overline{\textbf{M}}_t$. For every cell $i$ follows: \begin{multline}
\underbrace{m\big(M^i_{1:t}|\textbf{x}_{1:t}, \textbf{z}_{1:t}, \textbf{q}_{1:t})}_{m_{1:t}(M^i)}=\\\underbrace{m\big(M^i_{1:t-1}|\textbf{x}_{1:t-1}, \textbf{z}_{1:t-1}, \textbf{q}_{1:t-1}) }_{m_{1:t-1}(M^i)} \circledast 
\underbrace{m\big(\overline{M}^i_{t}|\textbf{x}_{t}, \textbf{z}_{t}, \textbf{q}_{t})}_{\overline{m}_{t}(M^i)}.
\label{MassUpdated_General} 
\end{multline} Thereby the combination rule $\circledast$ introduced by \cite{Tanzmeister2017} describes the handling of conflicting hypotheses. 

Before applying (\ref{MassUpdated_General}), $\textbf{M}_t$ receives one modification resulting in $\overline{\textbf{M}}_t$. The reason for this is the following: If measurements from previously unobserved areas or new objects in the environment occur, many unclassified particles exist in one cell. Consequently, a high evidence of unclassified occupancy ${m}_t(\{S,D\}^i)$ is derived. The original method handles the conflict between unclassified occupancy and other hypotheses with the use of Dempster's Rule (\ref{DS-Kombinationsregel}). This leads to artifacts of unclassified occupancy in the map, although avoiding these artifacts is one of the motivations for using DOGMs. Especially, radar measurements are noisier than lidar measurements, and ghost targets can occur in drivable areas. Therefore, error-prone detections also generate unclassified occupancy evidence masses. This work presents a solution by assigning the mass of unclassified occupancy $\overline{m}_t(\{S,D\}^i)$ to the unknown mass $\overline{m}_t(\Omega^i)$
\begin{equation}
\overline{m}_{t}(M^i)=\begin{cases}
m_{t}(\Omega^i)+m_{t}(\{S,D\}^i), & \; M^i=\Omega^i \\
0, & \; M^i=\{S,D\}^i \\
m_{t}(M^i), & \; \textrm{else}
\end{cases}.
\end{equation}  This increase of uncertainty over the map makes our approach more robust against noisy radar signals. 

\subsection{Clustering}
\label{Section:Clustering}
As a next step, the clustering of dynamic cells extracts object information from the resulting DOGM. Therefore, the first two statistical moments for the 2-D velocity of every grid cell are estimated as proposed in \cite{Nuss2017}. The algorithm only uses particles with age $a_{t,[k]}\geq a_{\textrm{min}}$ because newly generated ones would add noise to the estimation. This delay gives the tracking algorithm time for convergence before applying the clustering.
Dynamic cells are extracted from the set $I$ of all DOGM cells resulting in set  \begin{equation}
I_C=\{i \in I \: | \: m(D^i) \geq \epsilon_{D,\textrm{min}} \land m(D^i)>m(S^i) \}.	
\label{ClusteringEvidenzfiltering}
\end{equation}
Only cells of the DOGM with a dynamic mass $m(D^i)$ above a threshold $\epsilon_{D,\textrm{min}}$, and a higher dynamic than static mass $m(S^i)$ are considered resulting in more robustness against noise. 
After this, a clustering algorithm like in \cite{Steyer2017} and \cite{Gies2018} extracts dynamic cells. As a result, the method clusters dynamic cells that are close to each other and have similar velocity estimates. 

\section{Evaluation}
In this section, the developed radar-based DOGM approach is evaluated qualitatively and quantitatively. Therefore, it is compared to a lidar-based one in different urban environments and under different comparison metrics. 

\subsection{Experiment Configuration}
The method was implemented with C++ in our automated driving framework. A test vehicle, which is equipped with eleven Hella short-range radars and four Ibeo four-layer laser scanners, recorded real-world data. While the radar configuration covers \num{360}\textdegree\,  of the vehicle environment, the lidar configuration has blind spots on the sites. Due to the limited range of the used radar sensors, the grid has an edge length of \num{50}\,m, a resolution of $d_\textbf{G}=0{.}2$\,m, and a maximum number of particles $K^i_{\textrm{max}}=50$ per cell.

\subsection{Qualitative Evaluation}
\begin{figure} 
	\centering
	\resizebox{0.92\columnwidth}{!}
	{
		\begin{tikzpicture}
		\newcommand*\circled[1]{\tikz[baseline=(char.base)]{
				\node[shape=circle,draw,inner sep=0.5pt] (char) {#1};}}
		\setlength{\fboxsep}{0pt}%
		\setlength{\fboxrule}{0.8pt}%
		\begin{scope}[scale=1, xshift=-2.625cm, yshift=+3.35cm]
		\node {\fbox{\includegraphics[trim={10px 170px 18px 175px},clip,scale=0.358 ]{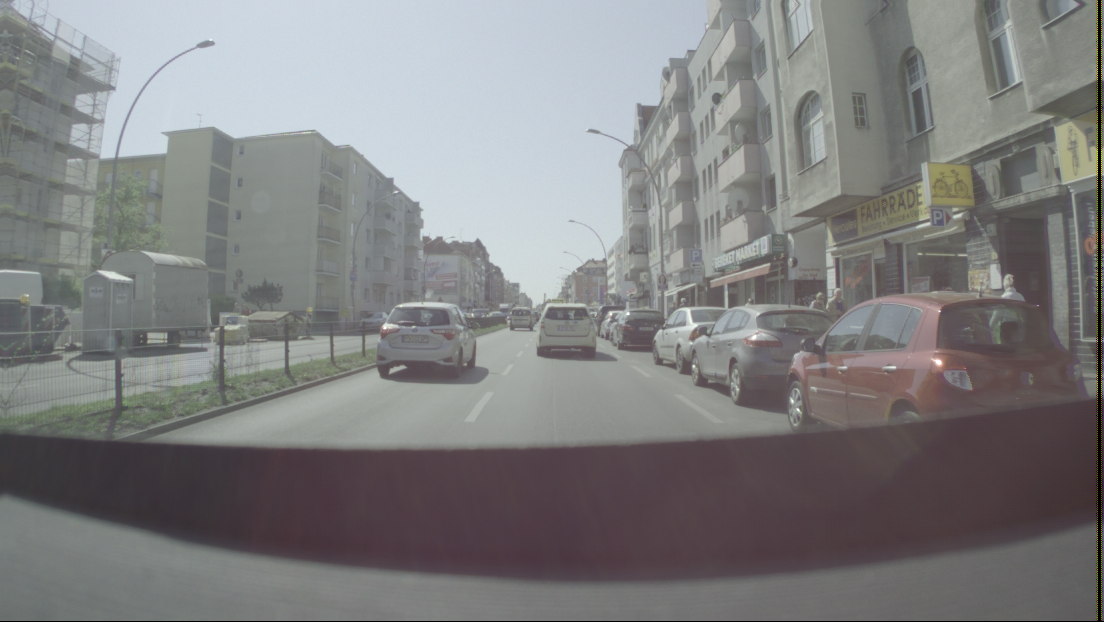}}};
		\node at(-1.2,-0.55) {$\circled{1}$};
		\node at(0.25,-0.55) {$\circled{2}$};
		
		\end{scope}
		
		\begin{scope}
		\node {\fbox{\includegraphics[scale=0.55]{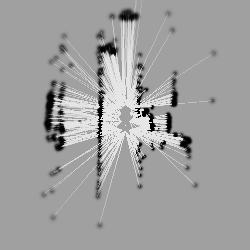}}};
		\node[opacity=1] at (0,0.133) {\includegraphics[scale=0.037]{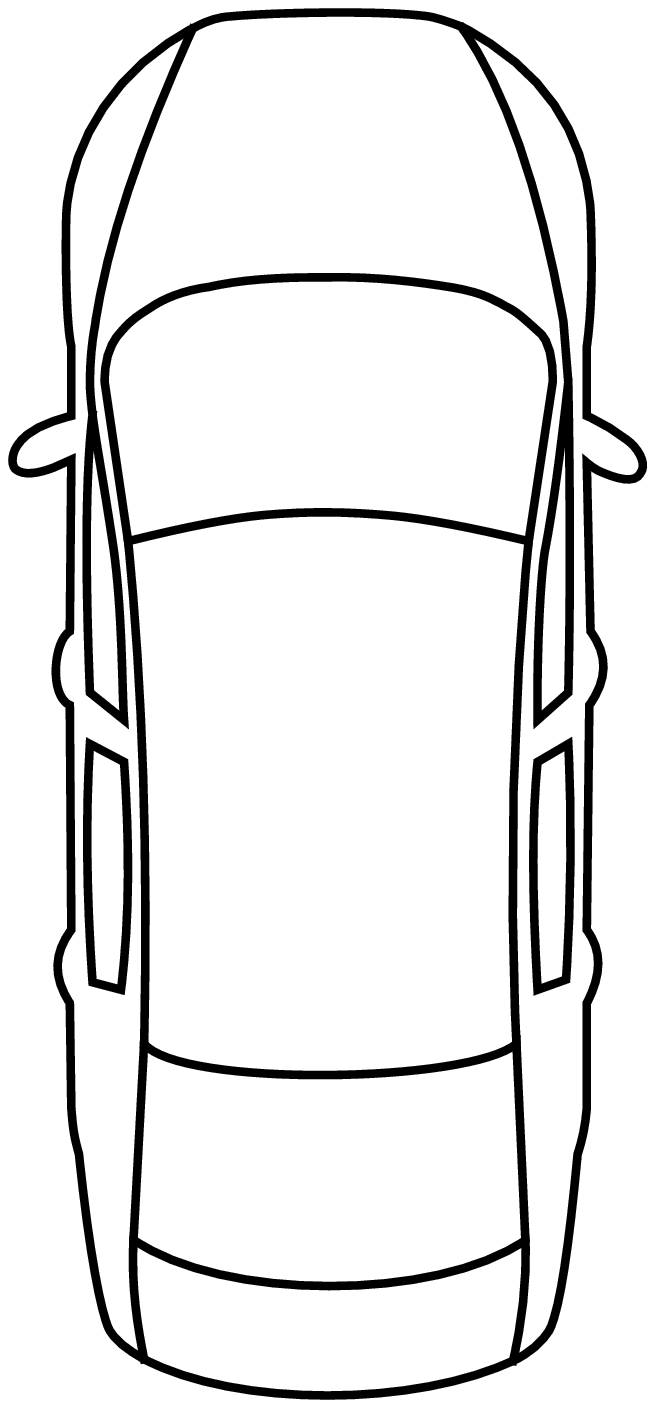}};
		\end{scope}
		
		\begin{scope}[xshift=-5.25cm]
		\node {\fbox{\includegraphics[scale=0.55]{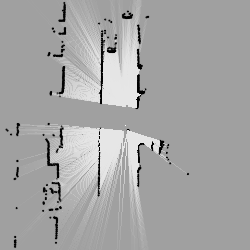}}};
		\node[opacity=1] at (0,0.133) {\includegraphics[scale=0.037]{car_topview_whitefilled_eps.eps}};
		\end{scope}

		\node[text width=5cm] at(-5.9,3.4) {(a)};
		
		\node[text width=5cm] at(-5.9,0) {(b)};
		\node[text width=5cm] at(-5.9,-5.25) {(c)};
		\node[text width=5cm] at(-5.9,-10.35) {(d)};
		\node[text width=5cm] at(-5.9,-14.2) {(e)};

		\begin{scope}[yshift=-5cm]
		\node {\fbox{\includegraphics[scale=0.55]{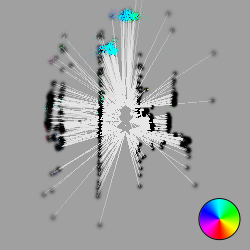}}};
		\node[opacity=1] at (1.835,-1.82){\includegraphics[scale=0.0467]{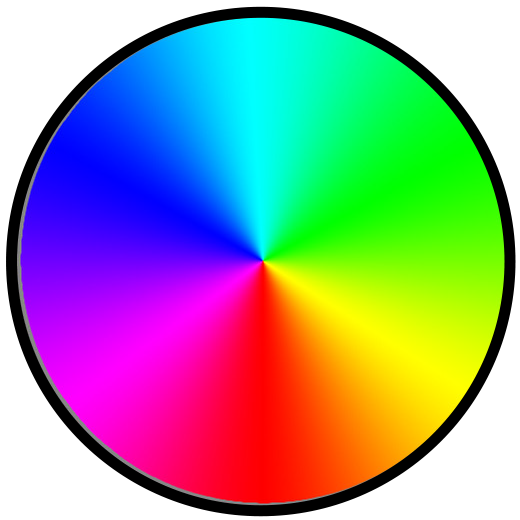}};
		\node[opacity=1] at (0,0.133) {\includegraphics[scale=0.037]{car_topview_whitefilled_eps.eps}};
		\node at(-0.2, 0.5) {$\circled{5}$};
		\node at(-1.7, 00.3) {$\circled{4}$};
		\end{scope}
		
		\begin{scope}[xshift=-5.25cm, yshift=-5cm]
		\node {\fbox{\includegraphics[scale=0.55]{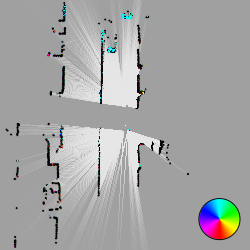}}};
		\node[opacity=1] at (1.835,-1.82){\includegraphics[scale=0.0467]{colorCircleRBA_Alpha_00000_00000.png}};
		\node[opacity=1] at (0,0.133) {\includegraphics[scale=0.037]{car_topview_whitefilled_eps.eps}};
		\node at(-1.5, 2.1) {$\circled{3}$};
		\end{scope}
		
		\begin{scope}[ yshift=-10cm]
		\node {\fbox{\includegraphics[scale=0.55]{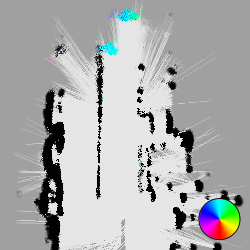}}};
		\node[opacity=1] at (1.835,-1.82){\includegraphics[scale=0.0467]{colorCircleRBA_Alpha_00000_00000.png}};
		\node[opacity=1] at (0,0.133) {\includegraphics[scale=0.037]{car_topview_whitefilled_eps.eps}};
		\node at(0, -0.6) {$\circled{6}$};
		\end{scope}
		
		\begin{scope}[xshift=-5.25cm, yshift=-10cm]
		\node {\fbox{\includegraphics[scale=0.55]{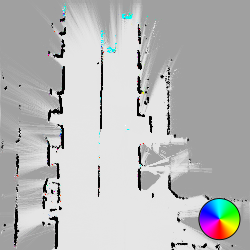}}};
		\node[opacity=1] at (1.835,-1.82){\includegraphics[scale=0.0467]{colorCircleRBA_Alpha_00000_00000.png}};
		\node[opacity=1] at (0,0.133) {\includegraphics[scale=0.037]{car_topview_whitefilled_eps.eps}};
		
		\end{scope}
		
		\begin{scope}[yshift=-14cm]
		\node {\fbox{\includegraphics[scale=0.55]{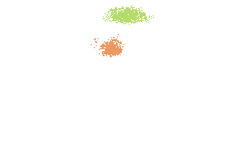}}};
		\node[opacity=1] at (0,-1.18) {\includegraphics[scale=0.037]{car_topview_whitefilled_eps.eps}};
		\end{scope}
		
		\begin{scope}[xshift=-5.25cm, yshift=-14cm]
		\node{\fbox{\includegraphics[scale=0.55]{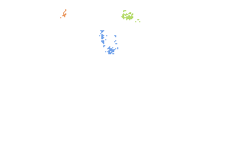}}};
		\node[opacity=1] at (0,-1.18) {\includegraphics[scale=0.037]{car_topview_whitefilled_eps.eps}};
		\end{scope}
		
		\end{tikzpicture}
	}
	\vspace{-0.3cm}
	\caption{Processing steps of the DOGM approach: (a) Camera image. (b) Evidential measurement grid $\textbf{M}_{t,z}$. (c) Classified evidential grid $\textbf{M}_{t}$. (d) DOGM $\textbf{G}_{1:t}$. (e) Clusters $C_t$.  Lidar-based (left) and radar-based (right) approach.}
	\label{VergleichAblaufSzenario1}
	\vspace{-0.5cm}
\end{figure}

The approach was evaluated qualitatively in different scenarios in Berlin's urban traffic. First, the ego-vehicle is driving straight ahead on an urban road. Two vehicles with approximately \num{50}\,km/h are driving in front. Parking vehicles and a metallic fence are located to the left and right as visualized in the camera image of Fig. \ref{VergleichAblaufSzenario1}\,(a).
Moreover, the processing steps of the lidar-based and the radar-based approach are illustrated in Fig. \ref{VergleichAblaufSzenario1}\,(b)-(e). 

The color coding is chosen as follows: Light grey visualizes the evidence of free space $m(F)$. Static occupancy evidence $m(S)$ and unclassified occupancy evidence $m(\{S,D\})$ are colored black. Unknown areas correspond to the dark grey color $m(\Omega)$. The dynamic evidence $m(D)$ is described with the illustrated color circle, whereas the color corresponds to the orientation of the estimated cell velocities. 
 For visualization purposes, a cell is visualized fully in the color of the color circle when the dynamic mass fulfills $m(D)>0.5$.

 The measurement grids of all sensors are fused using (\ref{DS-Kombinationsregel}), as shown in Fig. \ref{VergleichAblaufSzenario1}\,(b). Due to the higher angular resolution and lower noise behavior of lidar sensors are the contours of the environment better visible than in the radar-based DOGM. Fig. \ref{VergleichAblaufSzenario1}\,(c) illustrates the results of the grid-based tracking with subsequent evidence map estimation. Both methods correctly classify the dynamic objects $\large{\textcircled{\small{1}}}$ and $\large{\textcircled{\small{2}}}$ and determine the direction of the object's speed. However, due to the lacking capability of measuring velocities, the lidar-based approach also contains wrongly classified dynamic evidence in static areas, as shown in area $\large{\textcircled{\small{3}}}$. Here, particle velocities are sampled from a uniform distribution, and therefore also dynamic particles are generated in static areas. Because of the missing velocity information, dynamic particles have the same probability of surviving in a cell as static ones resulting in a high dynamic mass in newly observed areas. Additional radial velocity measurements reduce this effect. Nevertheless, the radar-based method also leads to errors due to noisy measurements caused by multi-path propagation as in areas of a metallic scaffold $\large{\textcircled{\small{4}}}$ or fence $\large{\textcircled{\small{5}}}$. In Fig. \ref{VergleichAblaufSzenario1}\,(d) visualizes the resulting DOGMs. It is observable that wrongly classified dynamic evidence is reduced due to temporal filtering of the maps, as described in equation (\ref{MassUpdated_General}). Another observable effect is the destruction of static evidence in area $\large{\textcircled{\small{6}}}$. This is due to the incorrect modeling of a direct line of sight in the radar-based inverse sensor model. Lastly, dynamic cells are clustered, as shown in Fig. \ref{VergleichAblaufSzenario1}\,(e) and both dynamic objects are successfully extracted. Because of the wrongly classified cells in the lidar-based method, an additional cluster is built in the static area $\large{\textcircled{\small{3}}}$. This underlines the added value of the additional speed measurement. 

\begin{figure}
	\centering
	\resizebox{\columnwidth}{!}
	{
		\begin{tikzpicture}
		\newcommand*\circled[1]{\tikz[baseline=(char.base)]{
				\node[shape=circle,draw,inner sep=0.5pt] (char) {#1};}}
		
		\setlength{\fboxsep}{0pt}%
		\setlength{\fboxrule}{0.8pt}%
		
		\begin{scope}[scale=1, xshift=-5.26cm, yshift=2.9cm]
		\node {\fbox{\includegraphics[trim={80px 80px 71px 53px},clip,scale=0.54 ]{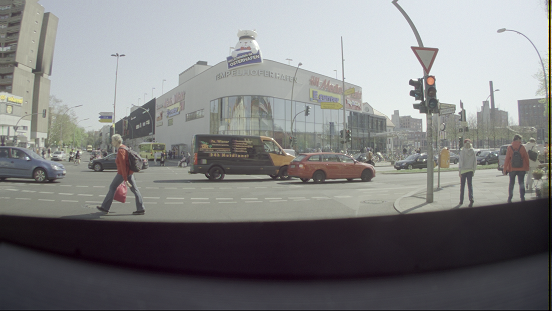}}};
		\end{scope}
		
		\begin{scope}[scale=1, xshift=0cm, yshift=2.9cm]
		\node {\fbox{\includegraphics[trim={80px 80px 71px 53px},clip,scale=0.54 ]{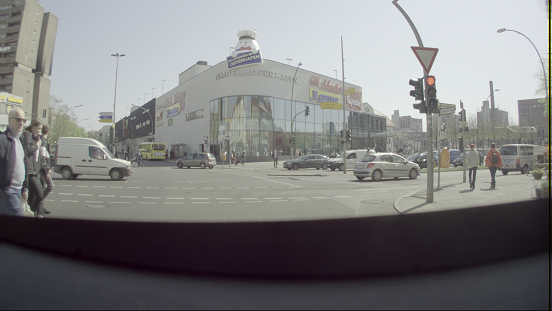}}};
		
		\end{scope}
		\begin{scope}[scale=1, xshift=+5.26cm, yshift=2.9cm]
		\node {\fbox{\includegraphics[trim={80px 80px 71px 53px},clip,scale=0.54 ]{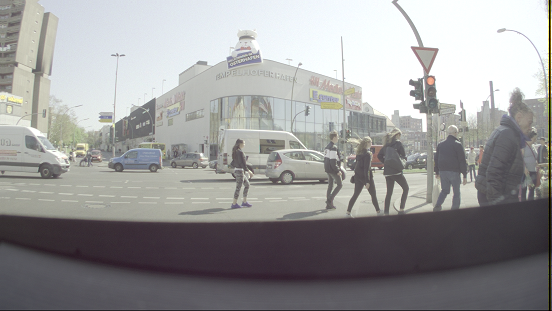}}};

		
		\end{scope}
		
		\begin{scope}[xshift=-5.26cm, yshift=0cm]
		\node{\fbox{\includegraphics[scale=0.71]{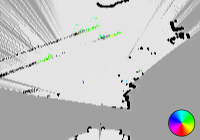}}};
		\node[opacity=1] at (0,-1.175) {\includegraphics[scale=0.043]{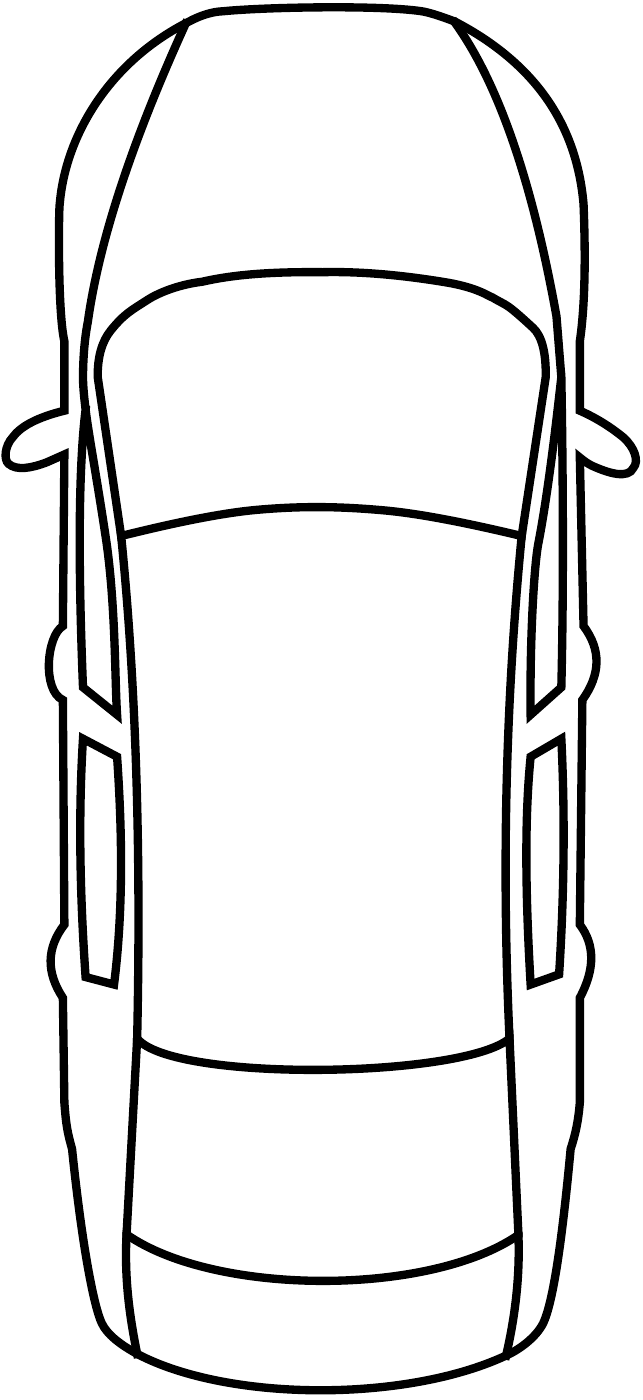}};
		\node[opacity=1] at (2.04,-1.32){\includegraphics[scale=0.04]{colorCircleRBA_Alpha_00000_00000.png}};
		\node at(-0.5,0.65) {$\large{\circled{1}}$};
		\node at(0,0.5) {$\large{\circled{2}}$};
		\node at(-0.5,-0.75) {$\large{\circled{3}}$};
		\node at(0.5,-0.5) {$\large{\circled{4}}$};
		\end{scope}
		
		\begin{scope}[xshift=0cm, yshift=0cm]
		\node{\fbox{\includegraphics[scale=0.71]{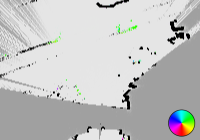}}};
		\node[opacity=1] at (0,-1.175) {\includegraphics[scale=0.043]{car_topview_whitefilled_eps}};
		\node[opacity=1] at (2.04,-1.32){\includegraphics[scale=0.04]{colorCircleRBA_Alpha_00000_00000.png}};
		\node at(-1.4,0.75) {$\large{\circled{5}}$};
		\end{scope}
		
		\begin{scope}[xshift=+5.26cm, yshift=0cm]
		\node{\fbox{\includegraphics[scale=0.71]{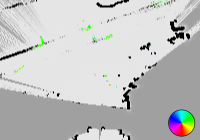}}};
		\node[opacity=1] at (0,-1.175) {\includegraphics[scale=0.043]{car_topview_whitefilled_eps}};
		\node[opacity=1] at (2.04,-1.32){\includegraphics[scale=0.04]{colorCircleRBA_Alpha_00000_00000.png}};
		\end{scope}
		
		\begin{scope}[xshift=-5.26cm, yshift=-3.7cm]
		\node{\fbox{\includegraphics[scale=0.71]{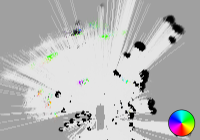}}};
		\node[opacity=1] at (0,-1.175) {\includegraphics[scale=0.043]{car_topview_whitefilled_eps}};
		\node[opacity=1] at (2.04,-1.32){\includegraphics[scale=0.04]{colorCircleRBA_Alpha_00000_00000.png}};
		\node at(-0.5,0.65) {$\large{\circled{1}}$};
		\node at(0,0.5) {$\large{\circled{2}}$};
		\node at(-0.5,-0.75) {$\large{\circled{3}}$};
		\node at(0.5,-0.5) {$\large{\circled{4}}$};
		\end{scope}

		\begin{scope}[xshift=0cm, yshift=-3.7cm]
		\node{\fbox{\includegraphics[scale=0.71]{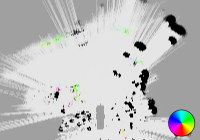}}};
		\node[opacity=1] at (0,-1.175) {\includegraphics[scale=0.043]{car_topview_whitefilled_eps}};
		\node[opacity=1] at (2.04,-1.32){\includegraphics[scale=0.04]{colorCircleRBA_Alpha_00000_00000.png}};
		\node at(-0.3,0.6) {$\large{\circled{6}}$};
		\node at(-1.4,0.75) {$\large{\circled{5}}$};
		\end{scope}
		
		\begin{scope}[xshift=+5.26cm, yshift=-3.7cm]
		\node{\fbox{\includegraphics[scale=0.71]{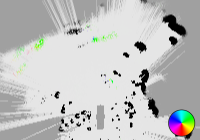}}};
		\node[opacity=1] at (0,-1.175) {\includegraphics[scale=0.043]{car_topview_whitefilled_eps}};
		\node[opacity=1] at (2.04,-1.32){\includegraphics[scale=0.04]{colorCircleRBA_Alpha_00000_00000.png}};
		\end{scope}
		
		\end{tikzpicture}
		
	}
	\vspace{-0.5cm}
	\caption{Comparison of DOGMs in an intersection scenario for different timestamps: First column: $t_1=0$\,s. Second column: $t_1=4$\,s. Third column: $t_1=8$\,s.  First row: Camera images. Second row: Results based on lidar measurements. Third row: Results based on radar measurements.}
	\label{Scenario2}
	\vspace{-0.5cm}
	
\end{figure}

In the second scenario, the ego-vehicle stands still at an urban intersection with crossing vehicles and pedestrians, as illustrated in the top row of Fig. \ref{Scenario2}. 
The increased challenges for the radar-based DOGM are threefold. First, the scenario includes more tangential traffic, while the radar measures radial velocities. Second, the vehicle is standing still, resulting in less relative velocity to the environment. Therefore, radar resolves static structures worse. Third, pedestrians have worse reflection properties than metallic cars.
Both methods detect and track the dynamic objects $\large{\textcircled{\small{1}}}$ - $\large{\textcircled{\small{4}}}$. However, the radar-based method has a partially wrong velocity estimation of the pedestrian $\large{\textcircled{\small{3}}}$ because of a purely tangential velocity component with respect to the sensor coordinate system. Therefore, the particle velocity component cannot be weighted by the measured radial velocity. Particles of both directions survive due to the high modeled occupancy uncertainty in the inverse sensor model of the measurement. In the lidar-based measurement grid, the standard deviation of the two-dimensional Gaussian is modeled much smaller such that only particles corresponding to the correct moving direction survive. The lidar-based method also has more artifacts of static evidence. These are caused by initializing static particles in the front of the vehicle. Then, they are validated through new measurements in the next timestamp and build a tail of static evidence when they reach the end of the vehicle. Accumulating static evidence with rule (\ref{MassUpdated_General}) leads to the observed behavior. The radar-based DOGM approach initializes and weights particles with the help of the measured radial velocity. This reduces the number of wrong static particles. Nevertheless, a wrong classification of static evidence in area $\large{\textcircled{\small{6}}}$  resulting from a wrong classification of the raw target is also observable. This can be caused when the raw target has a low relative velocity component in specific angular areas, especially when the ego-vehicle is stationary.

\subsection{Quantitative Evaluation}
This section evaluates the results for both sensor types using different metrics, which are also utilized in \cite{Nuss2017}. A real scenario is considered in which the ego-vehicle is stationary on a parking lot. A scenario without self-motion is chosen to investigate a behavior independent of odometry errors. Several static obstacles in the form of parked vehicles are located at the sites. In the front area, another vehicle approaches with a starting speed of about \num{20}\,km/h. Afterward, it performs a braking maneuver so that it stops in front of the ego-vehicle. \ref{QuanEval1}\,(a) visualizes the described scenario. 
\begin{figure*}[ht]
	\centering
	\resizebox{\textwidth}{!}
	{

}
	\vspace{-0.5cm}
	\caption{Quantitative evaluation: (a) Scenario overview (Image Source: Google Maps). (b) Estimated velocity of the moving object $\overline{v}_{x}^{L_\textrm{dyn}}$, $\large{\textcircled{\small{1}}}$ Delay $a_{\textrm{min}}$, $\large{\textcircled{\small{2}}}$ Movement without acceleration, $\large{\textcircled{\small{3}}}$ Braking maneuver, $\large{\textcircled{\small{4}}}$ Standstill. (c) Standard deviation $\sigma^{L_\textrm{dyn}}_{v_x}$. (d) Normalized estimation error squared $\eta$. (e) Receiver operating characteristic with a variation of the classification threshold $\epsilon_{\Lambda}$.}
	\label{QuanEval1}
	\vspace{-0.5cm}
\end{figure*}

\subsubsection{Velocity Estimation of a Moving Object}
This subsection evaluates the velocity estimation of the moving object. Therefore, grid cells corresponding with the moving vehicle are manually labeled for each point in time and are characterized by the set $L_\textrm{dyn}$. The mean velocity $\overline{v}_{x}^{L_\textrm{dyn}}$ of these cells in the $x$-direction of ego-vehicle 
 is estimated, and Fig. \ref{QuanEval1} shows the results of the experiment based on lidar and radar measurements. As a reference, we choose the outcome of the internal tracking algorithm of the Ibeo sensor setup. In the first interval $\large{\textcircled{\small{1}}}$, the estimation is delayed because only particles with a minimum age are used as described in section \ref{Section:Clustering}. Next, the vehicle does a non-accelerated movement in $\large{\textcircled{\small{2}}}$. The results of the radar method follow the reference, while the lidar-based one has a bias. This shift is due to the initialization of new particles from a distribution with zero mean. Therefore, static particles survive with a constant probability when the cells are validated through new lidar measurements in these cells. This tends the velocity estimation towards zero. In the radar-based method, Doppler velocity measurements can weight wrongly generated static particles. In interval $\large{\textcircled{\small{3}}}$, the moving object does a braking maneuver.  The radar-based velocity estimation again has a smaller deviation to the reference. However, some small outliers at $t=2{.}75$\,s and $t=3{.}05$\,s can be explained by the micro-Doppler effect. Our previous experiments showed that this effect often is caused by a rotating fan behind the radiator grille. Due to the additional velocity measurement, no new dynamic particles are initialized when the observed vehicle is stationary in interval $\large{\textcircled{\small{4}}}$. The lidar-based method, however, generates static and dynamic particles, and the latter can survive in static areas. Because of the stochastic properties of the used method, more particles with positive or more with a negative velocity component can survive, and the result is a bias in the estimation, as illustrated in interval $\large{\textcircled{\small{4}}}$. While the mean square error of the radar-based approach over the whole scenario is \num{0.138}\,m/s, the error of the lidar-based approach is \num{1.350}\,m/s, which is about \num{9.783} times higher.

\subsubsection{Consistency of the filter}
This section evaluates the consistency of the used filter. This is done by considering the normalized estimation error squared (NEES) $\eta$ with
\begin{equation} 
\eta=\frac{\big(\overline{v}^{L_\textrm{dyn}}_{x}-v^{L_\textrm{dyn}}_{x}\big)^2}{\big(\sigma^{L_\textrm{dyn}}_{v_x}\big)^2}, 
\end{equation} as described in \cite{Shalom2001}. The variable $v^{L_\textrm{dyn}}_{x}$ describes the reference $x$-velocity. Under the assumption of representing the vehicle as a Gaussian mixture, the combined variance $(\sigma_{v_x}^{L_\textrm{dyn}})^2$ is given by \begin{equation}
(\sigma_{v_x}^{L_\textrm{dyn}})^2=\sum_{i \in L_\textrm{dyn}}\frac{1}{|L_\textrm{dyn}|}\big( (\sigma_{v_x}^i)^2+(\overline{v}^i_{v_x})^2 \big) - \big(\overline{v}^{L_\textrm{dyn}}_{x}\big)^2
\end{equation} and Fig. \ref{QuanEval1}\,(c) illustrates the course of the combined standard deviation. The initial uncertainty in the radar-based approach is lower due to the additional velocity measurement. Fig. \ref{QuanEval1}\,(d) visualizes the results of the consistency test. The NEES values for lidar and radar measurements are compared with the one-sighted $95\%$ confidence interval. The lidar-based method exceeds the maximum value during the braking maneuver. This is due to the constant velocity modeling of the particle state transition. Therefore, the filter underestimates the uncertainty, and the estimation becomes inconsistent. Meanwhile, the radar-based estimation shows a smaller deviation from the reference, resulting in a consistent estimation throughout.

\subsubsection{Classification of Static and Dynamic Areas}
Now, the binary classification capability between static and dynamic areas is evaluated. Fig. \ref{QuanEval1}\,(e) visualizes the estimated Receiver Operating Curves under variation of the threshold $\epsilon_{\Lambda}\in [0,1]$ such that 
\begin{equation}
\Lambda=\begin{cases}
\{S\}, & \; m(S^i) \geq \epsilon_{\Lambda} \\
\{D\}, & \; \textrm{else}
\end{cases}
\end{equation} follows, as introduced in \cite{Steyer2018a}.
The static areas are associated with the positive class and dynamic areas vice versa. By comparing these two curves, the better classification performance of the radar-based approach resulting from the additional measured radial velocity is observable.
%
%
%

\section{Conclusion And Future Work}
This work has presented a radar-based approach for DOGM and object detection. An evidential grid-based tracking and mapping method has been extended for the use of only radar data, and a clustering algorithm extracts dynamic objects. This work presents as first an analysis of radar-based DOGM. Therefore, the approach has been evaluated and compared to a lidar-based method using real-world data collected in different urban environments. The resolution of objects in the latter is higher than in the former. However, the radar-based DOGM has more accurate and consistent velocity estimations and has better classification properties. Future work will fuse the results with semantic information from camera images to achieve more accurate results while maintaining a low-cost sensor suite. Furthermore, a sensor model, which better describes radar-specific wave propagation effects, should be used.
%
%
%
%
%
%
%
%
%
\bibliographystyle{IEEEtran}
\bibliography{IEEEabrv,ITSC_2020_ref}

\end{document}